\documentclass[onecolumn]{IEEEtran}

\IEEEoverridecommandlockouts
\usepackage{cite}
\usepackage{amsmath,amssymb,amsfonts}
\usepackage{algorithmic}
\usepackage{graphicx}
\usepackage{textcomp}
\usepackage{booktabs} 
\usepackage{xcolor}

\usepackage[font=footnotesize]{caption}
\def\BibTeX{{\rm B\kern-.05em{\sc i\kern-.025em b}\kern-.08em
    T\kern-.1667em\lower.7ex\hbox{E}\kern-.125emX}}

\begin{document}

\title{MOCHA: Discovering Multi-Order Dynamic Causality in Temporal Point Processes}

\author{%
\begin{tabular}{cccc}
\begin{tabular}[t]{c}
Yunyang Cao \\
\textit{Tongji University} \\
\end{tabular}
&
\begin{tabular}[t]{c}
Juekai Lin \\
\textit{Tongji University} \\
\end{tabular}
&
\begin{tabular}[t]{c}
Wenhao Li \\
\textit{Tongji University} \\
\end{tabular}
&
\begin{tabular}[t]{c}
Bo Jin \\
\textit{Tongji University} \\
\texttt{bjin@tongji.edu}
\end{tabular}
\end{tabular}
}

\maketitle

\begin{abstract}
Discovering complex causal dependencies in temporal point processes (TPPs) is critical for modeling real-world event sequences. Existing methods typically rely on static or first-order causal structures, overlooking the multi-order and time-varying nature of causal relationships. In this paper, we propose MOCHA, a novel framework for discovering multi-order dynamic causality in TPPs. MOCHA characterizes multi-order influences as multi-hop causal paths over a latent time-evolving graph. To model such dynamics, we introduce a time-varying directed acyclic graph (DAG) with learnable structural weights, where acyclicity and sparsity constraints are enforced to ensure structural validity. We design an end-to-end differentiable framework that jointly models causal discovery and TPP dynamics, enabling accurate event prediction and revealing interpretable structures. Extensive experiments on real-world datasets demonstrate that MOCHA not only achieves state-of-the-art performance in event prediction, but also reveals meaningful and interpretable causal structures.
\end{abstract}


\section{Introduction}
Event sequence data, where events occur asynchronously over time, is fundamental for modeling complex systems in domains such as clinical care \cite{schulam2017reliable}, finance \cite{bacry2015hawkes}, and recommender systems \cite{bonner2018causal}. Temporal point processes (TPPs) provide a powerful framework for modeling such event sequences, capturing temporal dependencies and predicting event timing and types \cite{hawkes1971spectra, zuo2020transformer, zhang2024neural}. However, most neural TPPs model only pairwise dependencies, ignoring multi-order causal structures.

Causal discovery in TPPs aims to reveal how events influence each other beyond surface-level correlations. Existing models typically capture only first-order direct influences \cite{du2016recurrent, mei2017neural, zhang2020self}, overlooking multi-order dependencies. Many models assume static structures \cite{zhang2020cause, gracious2023dynamic}, neglecting the temporal evolution of real-world systems. Furthermore, the absence of structural constraints often leads to black-box models with limited interpretability \cite{wang2023hierarchical,meng2024interpretable}.

\begin{figure}[ht]
    \centering
    \includegraphics[width=0.7\linewidth]{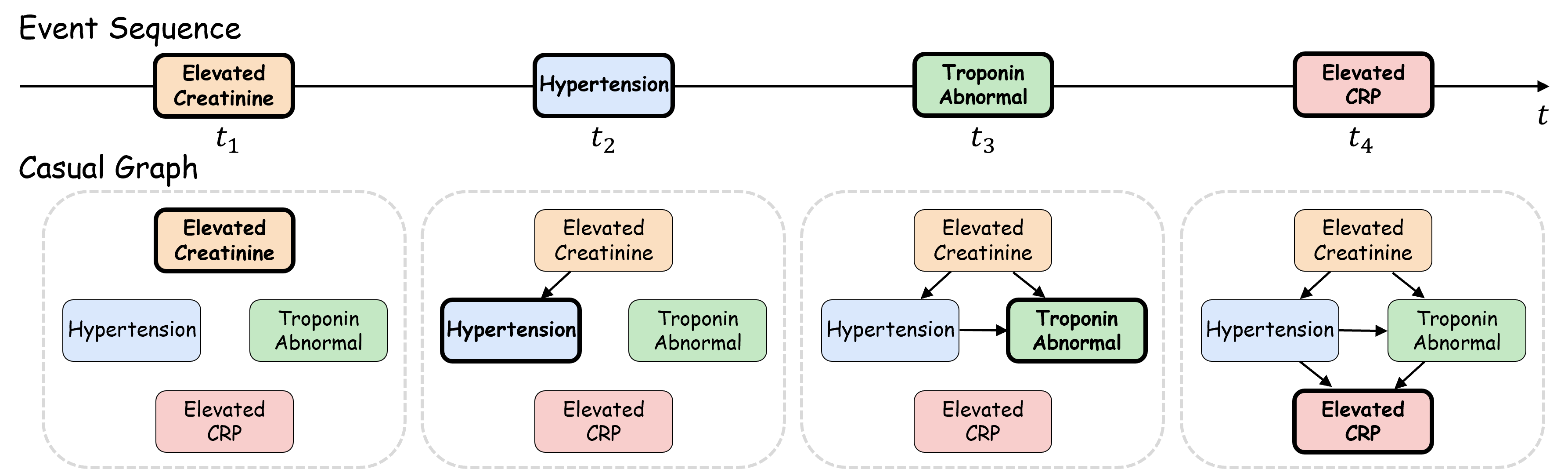}
    \caption{Illustration of an event sequence and its time-varying causal graphs in clinical settings. Multi-order paths capture indirect influences over time.}
    \label{fig:seq}
\end{figure}

The lack of multi-order and dynamic causal modeling presents critical limitations in representing the complexity of real-world event systems. 
For example, in clinical settings, an early sign of kidney dysfunction such as elevated creatinine may not only directly cause heart-related issues but also trigger intermediate conditions like hypertension, which eventually lead to abnormal cardiac troponin. Meanwhile, increased inflammation signals, such as elevated C-reactive protein (CRP), may be influenced by hypertension and abnormal troponin. Figure~\ref{fig:seq} shows a sample event sequence of this mechanism. Neglecting multi-order and converging dependencies leads to incomplete disease progression representations. Additionally, fixed causal structures fail to capture evolving factors such as patient conditions, treatment effects, and disease interactions. 

These limitations motivate a core research question: how to model multi-order dynamic causal dependencies. The primary challenge is the representation of multi-order causality, where direct influences alone are insufficient to reflect complex chains of influence. To address this, we characterize multi-order influences as multi-hop causal paths over a latent graph structure. Meanwhile, modeling temporal dynamics poses the difficulty of maintaining causal validity over time. We tackle this by introducing a time-varying directed acyclic graph (DAG), where acyclicity and sparsity constraints are enforced to ensure structural soundness. This formulation can decompose the complex problem of multi-order dynamic causality into a sequence of coupled static first-order structures, facilitating both interpretability and end-to-end learning.

To this end, we propose  \textit{Multi-Order Causal Hierarchical Architecture} (MOCHA), a unified framework for causal modeling in multivariate TPPs. MOCHA encodes event dependencies as a time-varying DAG, allowing multi-hop propagation while ensuring causal validity through acyclicity constraints. The contributions are as follows:
\begin{itemize}
    \item We represent multi-order causal dependencies into the intensity function via multi-hop paths over dynamic DAGs.
    \item  We design an end-to-end differentiable framework that unifies causal discovery with TPP modeling.
    \item Our model achieves superior likelihood performance and learns interpretable causal structures.
\end{itemize}

\section{Related Work}
\textbf{Temporal Point Processes (TPPs)} provide a probabilistic framework for modeling event sequences in continuous time. Classical models such as the Poisson process \cite{jorgenson1961multiple}, Hawkes process \cite{hawkes1971spectra}, and self-correcting process \cite{isham1979self} capture triggering influences through parametric excitation functions. Recent advances enhance their flexibility by incorporating deep neural networks. Neural TPPs encode historical events using RNNs \cite{du2016recurrent}, LSTMs \cite{mei2017neural}, or attention-based mechanisms \cite{zhang2020self, zuo2020transformer, yang2022transformer} to predict the next event. However, most rely on a Markovian assumption, modeling only short-term dependencies and failing to capture higher-order causal relationships across event sequences. To enhance interpretability, ITHP \cite{meng2024interpretable} introduces nonlinear functions to represent comprehensive event interactions. HCLTPP \cite{wang2023hierarchical} employs hierarchical contrastive learning to improve generalization. NJDTPP \cite{zhang2024neural} introduces a jump-diffusion framework to model non-stationary interactions. Despite these advances, existing models remain largely black-box, offering limited transparency into the underlying causal mechanisms.

\textbf{Causal discovery in temporal systems} can be broadly categorized into three main approaches.  
First, correlation-based and predictive causality methods, such as Granger causality \cite{granger1969investigating}, define causation in terms of improved forecasting. These include vector autoregressive models and their extensions to Hawkes processes \cite{xu2016learning}, as well as neural variants like CAUSE \cite{zhang2020cause}, which utilize attribution techniques in TPPs. However, such methods \cite{gracious2025neural} capture temporal correlations rather than genuine interventional or structural causality.
Second, counterfactual-based causality builds on frameworks like potential outcomes and treatment effect estimation. Rubin’s model \cite{rosenbaum1983central} is adapted to point processes for estimating average treatment effects \cite{gao2021causal}. Recent works identify key influencing events through counterfactual reasoning \cite{noorbakhsh2022counterfactual, zhang2022counterfactual}. Although effective for intervention analysis, these methods typically lack explicit modeling of multivariate or structural dependencies among event types.
Third, structural and graph-based causality aims to uncover interpretable causal graphs among events. Approaches include dynamic Bayesian networks \cite{spirtes2000causation, yang2024variational}, differentiable DAG learning \cite{zheng2018dags, yu2019dag, zhu2024causalnet}, and event-level discovery using sparse Hawkes matrices \cite{jalaldoust2022causal, qiao2023structural}. While more interpretable, these models often assume static or synchronized structures, limiting their ability to capture dynamic, multi-order causal dependencies in continuous-time event sequences.

\section{Preliminaries}
Our method builds upon the principles of multivariate TPPs and DAG structure learning. In this section, we provide essential background on these components. 

\textbf{Multivariate Temporal Point Processes.}
A multivariate temporal point process models a sequence of discrete events $\mathcal{S} = \{(t_i, k_i)\}_{i=1}^N$, where $t_i \in \mathbb{R}^+$ is the timestamp and $k_i \in \{1, \cdots, K\}$ is the type of the $i$-th event. The core concept in temporal point processes is the \textit{conditional intensity function}, which defines the instantaneous event rate for each event type $k$ given the history $\mathcal{H}_t$:
\begin{equation}
\lambda_k(t|\mathcal{H}_t) = \lim_{\Delta t \rightarrow 0} \frac{\mathbb{P}(\text{type } k \text{ occurs in } [t, t+\Delta t)|\mathcal{H}_t)}{\Delta t},
\end{equation}
where $\mathcal{H}_t$ denotes the history of all events up to time $t$. A representative example is the multivariate Hawkes process, where the intensity function is defined as:
\begin{equation} \label{eq:MTPP}
\lambda_k(t|\mathcal{H}_t) = \mu_k + \sum_{(t_j, k_j) \in \mathcal{H}_t} \phi_{k_j \rightarrow k}(t - t_j),
\end{equation}
where $\mu_k$ is the base intensity, $\phi_{k_j \rightarrow k}(\cdot)$ is a kernel function modeling the influence from event type $k_j$ to $k$. This formulation reflects the static influence of past events, serving as a foundation for causal reasoning in event sequences. 

\textbf{DAG Structure Learning.}
A directed acyclic graph (DAG) $\mathcal{G} = (\mathcal{V}, \mathcal{E})$ represents conditional dependencies among variables, where each node $v \in \mathcal{V}$ corresponds to a random variable (e.g., an event type), and each directed edge $(i \rightarrow j) \in \mathcal{E}$ indicates a direct causal influence from $i$ to $j$. The goal of causal discovery is to recover a DAG that faithfully reflects the underlying causal structure among variables \cite{pearl2009causality}, formulated as $\min_{\mathbf{W}} \mathcal{L}_{\text{fit}}(\mathbf{W}; \mathcal{D}) \quad \text{s.t.} \quad h(\mathbf{W}) = 0$, where $\mathbf{W} \in \mathbb{R}^{K \times K}$ is the weighted adjacency matrix of the DAG, $\mathcal{L}_{\text{fit}}$ is a loss measuring how well the DAG explains the data $\mathcal{D}$, $h(\mathbf{W})$ is an acyclicity constraint ensuring the learned graph is a DAG. A continuous formulation of the acyclicity constraint is introduced in NOTEARS \cite{zheng2018dags}:
\begin{equation} \label{eq:acy}
h(\mathbf{W}) := \text{Tr}(e^{\mathbf{W} \circ \mathbf{W}}) - K = 0,
\end{equation}
where $\circ$ is the Hadamard product, and $\text{Tr}$ is the trace. This formulation leverages the property that $\mathbf{W}$ corresponds to a DAG if and only if the trace of its matrix exponential equals the number of nodes.

\section{Methodology}
\begin{figure}[t]
    \centering
    \includegraphics[width=1.0\linewidth]{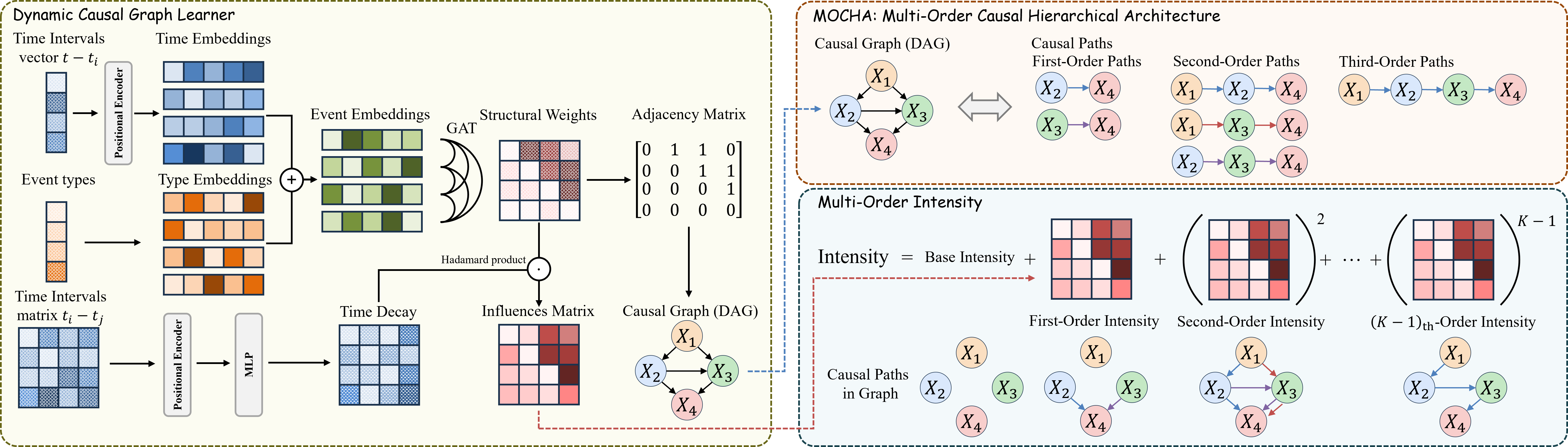}
    \caption{Overview of the proposed framework. The model consists of three key components. This figure illustrates the process from raw input data to multi-order causal structures and intensity functions at a fixed event timestamp. (1) Dynamic Causal Graph Learner, which encodes event types and time intervals to compute structural weights and temporal decay, and subsequently derives a causal graph (DAG) and an influence matrix representing event-to-event effects; (2) MOCHA (Multi-Order Causal Hierarchical Architecture), which decomposes the DAG into hierarchical causal paths including first-order and higher-order paths to capture multi-order causality; (3) Multi-Order Intensity Module, which models event intensities by aggregating base intensity and multi-order intensities derived from the multi-order causal paths. Together, the framework captures both structural and temporal dynamics of causal dependencies in complex event sequences.}
    \label{fig:stucture}
\end{figure}

We propose Multi-Order Causal Hierarchical Architecture (MOCHA), a neural framework that models multi-order and dynamic causality in multivariate temporal point processes.
Figure~\ref{fig:stucture} shows the framework of the MOCHA.

\subsection{Multi-Order Intensity Modeling}

The multi-order intensity is the sum of the first-order and higher-order intensity. To capture higher-order causal dependencies beyond direct excitation, we extend the classical formulation of the multivariate Hawkes process by recursively aggregating influence along multi-hop paths in a latent causal graph. In the standard multivariate Hawkes process shown in Eq.(\ref{eq:MTPP}), $\phi_{k_j \rightarrow k}(\cdot)$ represents the influence from event type $k_j$ to $k$, including both first-order and higher-order influences. We explicitly disentangle these influences by decomposing the function $\phi_{k_j \rightarrow k}(\cdot)$ into multi-order components. The first-order influence $\phi^{(1)}_{k_j \rightarrow k}(\cdot)$ corresponds to direct excitation, computed as the product of a structural weight and a temporal decay kernel:
\begin{equation} \label{eq:first_order}
\phi^{(1)}_{k_j \rightarrow k}(t - t_j) = w_{k_j \rightarrow k} \cdot \kappa(t - t_j),
\end{equation}
where $w_{k_j \rightarrow k}$ denotes the learned weight representing the direct causal influence from type $k_j$ to $k$, and $\kappa(\cdot)$ is a decay kernel, modeling how influence changes over time. In MOCHA, the weight $w_{k_j \rightarrow k}$ is time-varying, which can also be denoted as $w_{k_j \rightarrow k}(t)$.

Higher-order influences are modeled as propagating through multi-hop paths within a DAG $\mathcal{G}_t$. Specifically, for a path of length $l$ connecting event $(t_0, k_0) \rightarrow (t_1, k_1) \rightarrow \cdots \rightarrow (t_l, k_l)$, we define the total influence from an event $(t_0, k_0)$ to $(t_l, k_l)$ as the product of influence weights and decay kernels along the path:
\begin{equation} \label{eq:higher_order}
    \Phi^{(l)}_{\pi}(t - t_0) = \prod_{r=1}^{l} \left( w_{k_{r-1} \rightarrow k_r}(t_r - t_{r-1}) \cdot \kappa(t_r - t_{r-1}) \right),
\end{equation}
where $(t_l, k_l)=(t,k)$ is the last event in the path. $\pi = (k_0 \rightarrow k_1 \rightarrow \cdots \rightarrow k)$ is a length-$l$ causal path ending at type $k$. $w_{k_{r-1} \rightarrow k_r}(\cdot)$ is a dynamic edge weight in the DAG, which is learned from a neural network. $\kappa(\cdot)$ is a decay kernel, and different types of event share the same decay kernel. Based on the formulation of higher-order influences, we define the $l$-th order intensity for event type $k$ as:
\begin{equation}
    \lambda_k^{(l)}(t|\mathcal{H}_t) = \sum_{(t_i, k_i) \in \mathcal{H}_t} \sum_{\pi \in \mathcal{P}_l(k_i \rightarrow k)} \Phi^{(l)}_{\pi}(t - t_i),
\end{equation}
where $\mathcal{P}_l(k_i \rightarrow k)$ is the set of all causal paths of length $l$ from type $k_i$ to $k$ in DAG $\mathcal{G}_t$.

Therefore, we define the multi-order intensity for each type $k$ as the sum over all orders:
\begin{equation} \label{eq:intensity}
    \lambda_k(t|\mathcal{H}_t) = \mu_k + \sum_{l=1}^{L} \alpha_k^{(l)}\lambda_k^{(l)}(t|\mathcal{H}_t).
\end{equation}
where $L$ is the maximum order. $\mu_k$ is the base intensity of type $k$, which is a learnable parameter. $\lambda_k^{(l)}(t|\mathcal{H}_t)$ is the $l$-th order intensity. $\alpha_k^{(l)}$ is the learnable weight of the $l$-th order intensity for type $k$. Type $k\in\{1,2,\cdots,K\}$.

This multi-order formulation builds upon several modeling assumptions. First, the causal dependencies among event types are encoded in a DAG whose edge weights $w_{u \rightarrow v}$ are learnable. Second, the model considers only a finite number of causal hops ($\leq L$). Due to the directedness and acyclicity of DAG, the maximum length of the causal path is less than the total number of nodes $K$, i.e., $L<=K-1$. Third, intermediate events act as carriers of causal signals, enabling the recursive transmission of influence and facilitating higher-order interactions across the event sequence. The intensity function of MOCHA generalizes the Hawkes process and enables interpretable modeling of structured, temporally-evolving causal influences. 

Dynamic influence refers to the adaptive, time-varying relationships between events. MOCHA dynamically constructs influence functions by conditioning both the structural weights and the time decay on contextual event features. This allows the model to adapt the influence strength and decay based on time and event history.

\subsubsection{Structural Weights}  
The structural weights at time $t$ are denoted as a matrix $\mathbf{W}_t \in \mathbb{R}^{K \times K}$, where each entry $\mathbf{W}_t[u, v] = w_{u \rightarrow v}(t)$ captures the dynamic causal influence from event type $u$ to $v$ at time $t$. These weights are learned through a Graph Attention Network (GAT) \cite{velivckovic2018graph} that dynamically adjusts the influence based on the event history and the specific event types involved. 

Let the last occurrence times of all event types be represented as $\mathbf{t}^{last} = (t_1^{last}, t_2^{last}, \cdots, t_K^{last})^T$, where $K$ is the total number of event types. Then, the time interval at time $t$ is denoted as $\boldsymbol{\tau}=(\tau_1, \tau_2, \cdots, \tau_K)$, where $\tau_k=t-t_k^{last}$. To incorporate temporal context into the structural weights, we encode the time intervals $\boldsymbol{\tau}$ using a positional encoding function $\mathrm{PE}: \mathbb{R} \rightarrow \mathbb{R}^{2d}$ inspired by the sinusoidal encoding in Transformer architectures. Specifically, for each dimension $i=1,\cdots,d$, the $i$-th and $(i+d)$-th components of $\mathrm{PE}(\tau_k)$ are defined as:
\begin{equation}
\begin{split}
    \mathrm{PE}(\tau_k)[i] &= \sin\left(\tau_k / 10000^{i/d}\right), \\
    \mathrm{PE}(\tau_k)[i+d] &= \cos\left(\tau_k / 10000^{i/d}\right),
\end{split}
\end{equation}
yielding a $2d$-dimensional temporal embedding vector for each $\tau_k$. We denote the matrix of all temporal embeddings as $\mathbf{E}_{\text{time}} \in \mathbb{R}^{K \times 2d}$, where the $k$-th row corresponds to $\mathrm{PE}(\tau_k)$. To incorporate type information into the structural weights, each event type $k \in \{1, \dots, K\}$ is associated with a learnable type embedding vector $\mathbf{e}^{\text{type}}_k \in \mathbb{R}^{2d}$, and we stack them into a matrix $\mathbf{E}_{\text{type}} \in \mathbb{R}^{K \times 2d}$. Hence, we obtain the combined dynamic embedding for all event types as
\begin{equation}
    \mathbf{H}_t = \mathbf{E}_{\text{time}} + \mathbf{E}_{\text{type}} \in \mathbb{R}^{K \times 2d}.
\end{equation}
These embeddings incorporate both temporal recency and event-type semantics into the structural reasoning process.

To capture the dynamic influence among event types, we apply a GAT over the combined embedding matrix $\mathbf{H}_t \in \mathbb{R}^{K \times 2d}$. The attention mechanism learns the pairwise structural weights based on query-key interactions between event types. Specifically, for each pair $(u,v)$ of event types, we compute the attention score $e_{u \rightarrow v}(t)$ as:
\begin{equation}
    e_{u \rightarrow v}(t) = \text{LeakyReLU}\left( \mathbf{a}^\top [\mathbf{W}_Q \mathbf{h}_u(t) \,\|\, \mathbf{W}_K \mathbf{h}_v(t)] \right),
\end{equation}
where $\mathbf{h}_u(t), \mathbf{h}_v(t) \in \mathbb{R}^{2d}$ are the $u$-th and $v$-th rows of $\mathbf{H}_t$, respectively, $\mathbf{W}_Q, \mathbf{W}_K \in \mathbb{R}^{d' \times 2d}$ are learnable projection matrices for query and key, $\mathbf{a} \in \mathbb{R}^{2d'}$ is the attention vector, and $\|$ denotes concatenation. These attention scores are first aggregated into attention embeddings by computing a context vector for each event type $v$:
\begin{equation}
    \tilde{\mathbf{h}}_v(t) = \sum_{u=1}^{K} \eta_{u \rightarrow v}(t) \cdot \mathbf{h}_u(t),
\end{equation}
where $\eta_{u \rightarrow v}(t)$ is the normalized attention weight obtained by $\eta_{u \rightarrow v}(t) = \exp(e_{u \rightarrow v}(t)) \big/ \sum_{u'=1}^{K} \exp(e_{u' \rightarrow v}(t))$.
The dynamic structural weights are then computed by projecting each context vector $\tilde{\mathbf{h}}_v(t)$ into a $K$-dimensional weight vector using a linear transformation:
\begin{equation}
    \mathbf{w}_v(t) = \mathbf{W}_{\text{proj}} \tilde{\mathbf{h}}_v(t),
\end{equation}
where $\mathbf{W}_{\text{proj}} \in \mathbb{R}^{K \times 2d}$ is a learnable weight matrix. Stacking the resulting vectors $\{\mathbf{w}_v(t)\}_{v=1}^K$ row-wise gives the final dynamic structural weight matrix:
\begin{equation}
    \mathbf{W}_t = \left[\mathbf{w}_1(t)^T; \cdots; \mathbf{w}_K(t)^T\right] \in \mathbb{R}^{K \times K},
\end{equation}
where each entry $\mathbf{W}_t[u, v]$ reflects the structured causal strength from type $u$ to $v$ at time $t$.

\subsubsection{Time Decay}
Effective temporal modeling requires capturing how past influences fade over time. To this end, we introduce a learnable time decay module that maps the elapsed time since past events to a scalar attenuation factor. Unlike classical exponential kernels with static decay rates, our formulation enables context-aware and non-parametric decay patterns, conditioned on event timing.

Let $\Delta t>0$ denote the elapsed time since a past event occurred. We first apply a sinusoidal positional encoding $\mathrm{PE}: \mathbb{R} \rightarrow \mathbb{R}^{2d}$ to transform the scalar $\Delta t$ into a $2d$-dimensional embedding $\mathbf{e}_{\Delta t} = \mathrm{PE}(\Delta t) \in \mathbb{R}^{2d}$. Then, we pass this embedding through a lightweight multi-layer perceptron (MLP) to produce a scalar decay score:
\begin{equation}
    \kappa(\Delta t) = \text{Sigmoid}(W_{FC_2}\text{ReLU}(W_{FC_1}\mathbf{e}_{\Delta t}+b_1)+b_2),
\end{equation}
where $W_{FC_1}, W_{FC_2}$ and biases $b_1,b_2$ are learnable parameters of a two-layer MLP. To ensure numerical stability, the decay is explicitly set to zero: $\kappa(\Delta t) = 0$ when $\Delta t \leq 0$. The Sigmoid function controls that $0<\kappa(\Delta t)<1$.

Finally, the final influence of a past event of type $u$ at time $t_i$ on type $v$ at time $t$ is computed as:
\begin{equation}
    \phi_{u \rightarrow v}(t - t_i; t) = w_{u \rightarrow v}(t) \cdot \kappa(t - t_i),
\end{equation}
where $w_{u \rightarrow v}(t)$ is the structural weight, and $\kappa(t - t_i)$ captures the dynamic time decay described in Eq.(\ref{eq:first_order}-\ref{eq:higher_order}). 

Together, the structural weights and time decay functions define a dynamic influence function that is both context-aware and temporally adaptive. The next section introduces how these structural weights are embedded into a learnable causal graph structure.

\subsection{Dynamic Causal Graph Learning}
MOCHA learns time-varying DAGs to capture dynamic causal structures more accurately. The dynamic structural weight matrix $\mathbf{W}_t \in \mathbb{R}^{K \times K}$ characterizes the causal influence among event types at time $t$. To interpret and utilize these dynamic influences in a graph-theoretic manner, we construct a time-evolving DAG $\mathcal{G}_t = (\mathcal{V}, \mathcal{E}_t)$, where $\mathcal{V} = \{1, 2, \dots, K\}$ denotes the set of event types, and $\mathcal{E}_t$ represents the set of directed edges indicating the learned causal relations at time $t$. Each edge $(u \rightarrow v) \in \mathcal{E}_t$ exists if and only if the corresponding weight $\sigma(\mathbf{W}_t[u, v])$ exceeds a threshold $\theta$. Formally, the adjacency matrix $\mathbf{A}_t \in \{0,1\}^{K \times K}$ of the graph $\mathcal{G}_t$ is defined as:
\begin{equation}
    \mathbf{A}_t[u, v] = 
    \begin{cases}
        1, & \text{if } \sigma(\mathbf{W}_t[u, v]) > \theta, \\
        0, & \text{otherwise}.
    \end{cases}
\end{equation}
The activation function is defined as $\sigma(w) = 1 - \exp(-\beta |w|)$, where $|w|$ is the absolute value of $w$, and $\beta>0$ and $\theta\geq0$ are fixed hyperparameters. The non-linear activation $\sigma(w)$ ensures that small weights are suppressed while preserving significant causal influences.

\subsubsection{Causality}
Causality is formally defined through the lens of Structural Causal Models (SCMs) \cite{pearl2009causality}, which model each variable as a function of its parents in a causal graph. In MOCHA, variables represents event types, and their parents are all variables pointing to them through the edges $\mathcal{E}_t$ in the DAG $\mathcal{G}_t$. Hence, the hypotheses of causality in MOCHA are as follows. If $(u \rightarrow v) \in \mathcal{E}_t$ (i.e., $A[u,v] \neq 0$), then $u$ is considered a potential cause of $v$ under the learned structural assumptions. Otherwise, if $A[u,v]=0$, then $u$ is not the cause of $v$. 

Higher-order causality can be represented as the powers of adjacency matrix $\mathbf{A}_t$. Specifically, the entry $[\mathbf{A}_t^l]_{u,v}$ counts the number of directed paths of length $l$ from event type $u$ to $v$ at time $t$. Since $\mathcal{G}_t$ is a DAG, it contains no cycles and the longest directed path has length at most $K - 1$. Therefore, $\mathbf{A}_t^K = \mathbf{0}$, where $\mathbf{0}$ is the zero matrix. This property ensures that all higher-order causal dependencies are finitely bounded, so we can set the maximum order $L=K-1$.

\subsubsection{Acyclicity}
To interpret the proposed dynamic structural weights in a causal framework, we adopt a continuous and differentiable DAG constraint inspired by Eq.(\ref{eq:acy}). Specifically, we treat $|\mathbf{W}_t|$ as a soft adjacency matrix of the causal graph at time $t$, and use the following acyclic constraint:
\begin{equation}
    h(\mathbf{W}_t) = \text{Tr}(e^{|\mathbf{W}_t| \circ |\mathbf{W}_t|}) - K,
\end{equation}
where $\text{Tr}(\cdot)$ is the trace function and the trace of $\exp(|\mathbf{W}_t| \circ |\mathbf{W}_t|)$ counts the total weight of all cycles in the graph, so enforcing $h(\mathbf{W}_t)=0$ guarantees acyclicity. Furthermore, for a sequence $\mathcal{S} = \{(t_i, k_i)\}_{i=1}^N$, we incorporate it into the training objective as a soft regularization term:
\begin{equation} \label{eq:acyc}
    \mathcal{L}_{\text{acyclic}} = \gamma_{\text{acyclic}} \sum_{i=1}^{N}{h(\mathbf{W}_{t_i})},
\end{equation}
where $\gamma_{\text{acyclic}}$ is a tunable coefficient. This design ensures that MOCHA learns dynamic structural weights and causal graph topologies in an end-to-end differentiable manner.

\subsubsection{Sparsity}
Causal graphs in real-world event systems are typically sparse, with each event type influenced by only a few others. To reflect this prior and improve interpretability, we encourage sparsity in the structural weights $\mathbf{W}_t$ via an $\ell_1$ regularization term:
\begin{equation} \label{eq:spar}
    \mathcal{L}_{\text{sparse}} = \gamma_{\text{sparse}} \sum_{i=1}^{N} \|\mathbf{W}_{t_i}\|_1,
\end{equation}
where $\|\cdot\|_1$ denotes the element-wise $\ell_1$ norm, and $\gamma_{\text{sparse}}$ is a regularization coefficient. This promotes a compact dynamic causal graph by shrinking insignificant edges toward zero.

Overall, the dynamically constructed graph $\mathcal{G}_t$ enables MOCHA to reason about causal structure in a temporally adaptive, data-driven fashion. By grounding influence functions in an explicit graph, the model supports both predictive accuracy and interpretability in complex temporal systems.

\subsection{Training and Inference}
MOCHA is trained to model the conditional intensity function $\lambda_k(t)$ for each event type $k \in \{1, \dots, K\}$, enabling likelihood-based learning of temporal point processes. Given an event sequence $\mathcal{S} = \{(t_i, k_i)\}_{i=1}^{N}$ observed in $[0, T]$, the negative log-likelihood objective is:
\begin{equation}
    \mathcal{L}_{\text{NLL}} = - \sum_{i=1}^{N} \log \lambda_{k_i}(t_i|\mathcal{H}_{t_i}) + \sum_{k=1}^{K} \int_{0}^{T} \lambda_k(t|\mathcal{H}_t) dt.
\end{equation}
The first term encourages high intensity at the true event times, while the second term penalizes overprediction by integrating the predicted intensity over time. We compute $\lambda_k(t|\mathcal{H}_t)$ using the Eq.(\ref{eq:intensity}). The integral is approximated using the trapezoidal rule over discretized time steps. To enable causal interpretability and enforce structural assumptions, we incorporate additional regularization terms for acyclicity and sparsity, resulting in the full training objective:
\begin{equation}
    \mathcal{L} = \mathcal{L}_{\text{NLL}} + \mathcal{L}_{\text{acyclic}} + \mathcal{L}_{\text{sparse}},
\end{equation}
where the latter two terms are defined in Eq.(\ref{eq:acyc}-\ref{eq:spar}). The entire model is trained end-to-end using gradient-based optimization.

During inference, MOCHA computes the intensity function $\lambda_k(t)$ by dynamically generating $\mathbf{W}_t$ and $\kappa(\Delta t)$ based on the event history $\mathcal{H}_t$. This enables efficient simulation, forecasting, and causal analysis of future events under complex and evolving dynamics.

\section{Experiments}
\noindent \textbf{Experimental Setup}

\textbf{Datasets.}
We use seven real-world datasets: \textit{AKI} (Acute Kidney Injury), \textit{CAD} (Coronary Artery Disease), \textit{SE} (Sepsis), \textit{ST} (Stroke), \textit{RT} (Retweet) \cite{zhou2013learning}, \textit{TB} (Taobao) \cite{xue2022hypro} and \textit{SO} (StackOverflow) \cite{leskovec2016snap}. AKI, CAD, SE and ST are medical datasets extracted from MIMIC-IV \cite{johnson2018mimic}. 

\textbf{Baselines.}
We compare \textit{MOCHA} with a variety of TPP models: \textit{RMTPP} \cite{du2016recurrent}, \textit{NHP} \cite{mei2017neural}, \textit{SAHP} \cite{zhang2020self}, \textit{THP} \cite{zuo2020transformer}, \textit{AttNHP} \cite{yang2022transformer}, \textit{HCLTPP} \cite{wang2023hierarchical}, \textit{ITHP} \cite{meng2024interpretable}, \textit{NJDTPP} \cite{zhang2024neural}. The implementation also refers to EasyTPP \cite{xue2024easytpp}.

\textbf{Evaluation Metrics.}
We evaluate model performance from both predictive and causal perspectives. First, we calculate the \textit{Negative Log-Likelihood (NLL)} to assess the quality of the TPP modeling. Second, we measure the \textit{Root Mean Squared Error (RMSE)} of event time prediction and the \textit{accuracy} of event type prediction, shown in the supplementary materials. Third, we compute the causal path \textit{matching rate} to assess the model’s ability to recover known causal relations from ground-truth domain knowledge. Fourth, we provide \textit{qualitative visualizations} of the learned dynamic causal graphs to assess interpretability. 

These evaluation metrics are designed to answer four research questions (RQ).

\vspace{\baselineskip}

\noindent \textbf{RQ1: Can Modeling Multi-Order Dynamics Improve Performance?}

To evaluate the effectiveness of modeling multi-order dynamic causal dependencies, we compare MOCHA with state-of-the-art TPP models on a diverse set of clinical and social datasets. As shown in Table~\ref{Main_NLL_conpact}, MOCHA consistently achieves the lowest NLL across all seven datasets, indicating an improved capacity to model complex event sequences.

\begin{table}[ht]
\setlength{\tabcolsep}{1.5mm}
\centering
\caption{Negative Log-Likelihood (NLL) comparison across benchmarks ($\downarrow$ is better).}
\begin{tabular}{lccccccc}
\toprule
& AKI & CAD & SE & ST & RT & TB& SO\\
\midrule
RMTPP &5.40&3.26&2.24&2.89&4.42&1.48&2.88 \\
NHP &4.42&2.78&1.44&1.98&4.12&0.74&2.67 \\
SAHP &4.55&2.33&2.63&1.04&4.16&0.76&3.51 \\
THP & 4.89&2.06&1.38&1.37&4.54&0.86&2.40 \\
AttNHP & 5.07&1.81&2.21&2.08&4.27&1.19&2.65 \\
HCLTPP &5.03&1.99&1.52&2.08&3.58&1.50&2.87 \\
ITHP &4.39&1.96&1.59&0.92&4.22&0.97&3.23 \\ 
NJDTPP &3.92&2.95&2.38&3.02&4.09&1.39&2.39 \\
MOCHA &\textbf{3.80}&\textbf{0.69}&\textbf{0.70}&\textbf{0.91}&\textbf{3.51}&\textbf{-0.36}&\textbf{2.32} \\
\bottomrule
\end{tabular}
\label{Main_NLL_conpact}
\end{table}

The performance gain is especially pronounced on clinical datasets (AKI, CAD, SE, ST), where long-range dependencies and evolving interactions are common. For instance, MOCHA achieves an NLL of 0.69 on CAD and 0.70 on SE, outperforming the best baselines (ITHP and THP) by wide margins. These results suggest that MOCHA is better at leveraging rich and temporally extended medical histories to estimate the likelihood of future clinical events.

These results confirm that explicitly modeling multi-order dynamic causality can enhance the performance of model fitting. By jointly learning structural weights and adaptive time decay, MOCHA enhances its ability to model complex temporal dependencies and evolving causal structures.

\vspace{41pt}

\noindent \textbf{RQ2: How Do Multi-Order and Dynamic Structures Influence Performance?}

To disentangle the impact of multi-order and dynamic structures in MOCHA, we conduct an ablation study by incrementally adding key modules to a basic Hawkes process model. The results are summarized in Table~\ref{tab:ablation}. 

\begin{table}[ht]
\centering
\caption{Ablation study of MOCHA across seven datasets. The leftmost column shows model variants obtained by incrementally adding key modules: starting from a univariate Hawkes process, then incorporating multivariate structure, multi-order causality, and finally the dynamic causal graph. Lower NLL indicates better performance.}
\setlength{\tabcolsep}{1mm}
\begin{tabular}{lccccccc}
\toprule
Model Variant & AKI & CAD & SE & ST & RT & TB & SO \\
\midrule
Hawkes & 5.88 & 2.17 & 3.06 & 3.14 & 4.67 & 1.31 & 5.38 \\
+ Multivariate & 5.73 & 2.03 & 2.77 & 2.86 & 4.62 & 0.04 & 3.81 \\
+ Multi-Order & 5.71 & 1.66 & 2.16 & 2.27 & 4.41 & -0.02 & 3.64 \\
+ Dynamic & \textbf{3.80} & \textbf{0.69} & \textbf{0.70} & \textbf{0.91} & \textbf{3.51} & \textbf{-0.36} & \textbf{2.32} \\
\bottomrule
\end{tabular}
\label{tab:ablation}
\end{table}

Starting from a univariate Hawkes process, we observe that modeling only self-excitation is insufficient for complex data. Extending to a multivariate Hawkes process captures inter-type interactions, yielding modest improvements. Introducing multi-order causal modeling substantially reduces NLL, particularly in datasets like SE and ST, where long-range, indirect influences are common. Finally, incorporating the dynamic causal graph learning mechanism, MOCHA achieves the best performance across all datasets. The explicit modeling of time-varying causality yields the largest improvements in NLL, particularly on clinical datasets, where event patterns are known to change over time.

These results indicate that multi-order structures enable richer modeling of indirect dependencies, while dynamic modeling captures temporally evolving causality, both of which are essential for accurate event prediction. The full MOCHA model benefits from the combined effect of these components, achieving consistent improvements across all benchmarks.

\vspace{\baselineskip}

\noindent \textbf{RQ3: Can MOCHA Recover Ground-Truth Causal Paths?}

To evaluate the validity of the learned dynamic causal graphs, we examine whether MOCHA can recover ground truth paths, such as medically known causal paths related to Acute Kidney Injury (AKI) Phase III. Based on authoritative clinical guidelines \cite{kdigo2012kdigo} and previous research \cite{pan2022predialysis, zhou2022lactate, chen2024prognostic}, we conclude 10 representative causal paths leading to AKI Phase III, including 4 first-order paths and 6 higher-order paths. We match these paths against the learned dynamic DAGs at the time when Phase III events occur. A causal path is considered matched if all intermediate and terminal directed edges comprising the path appear in the inferred DAG $\mathcal{G}_t$ at the relevant time. This strict criterion ensures alignment with the full causal path rather than partial overlap. For each path, we compute its matching rate. Some causal paths are only relevant to specific patient subgroups, so the theoretical maximum matching rate is inherently below 100\%. Generally, a higher matching rate indicates better alignment between the model's inferred causality and established knowledge.

\begin{figure}[ht]
    \centering
    \includegraphics[width=0.5\linewidth]{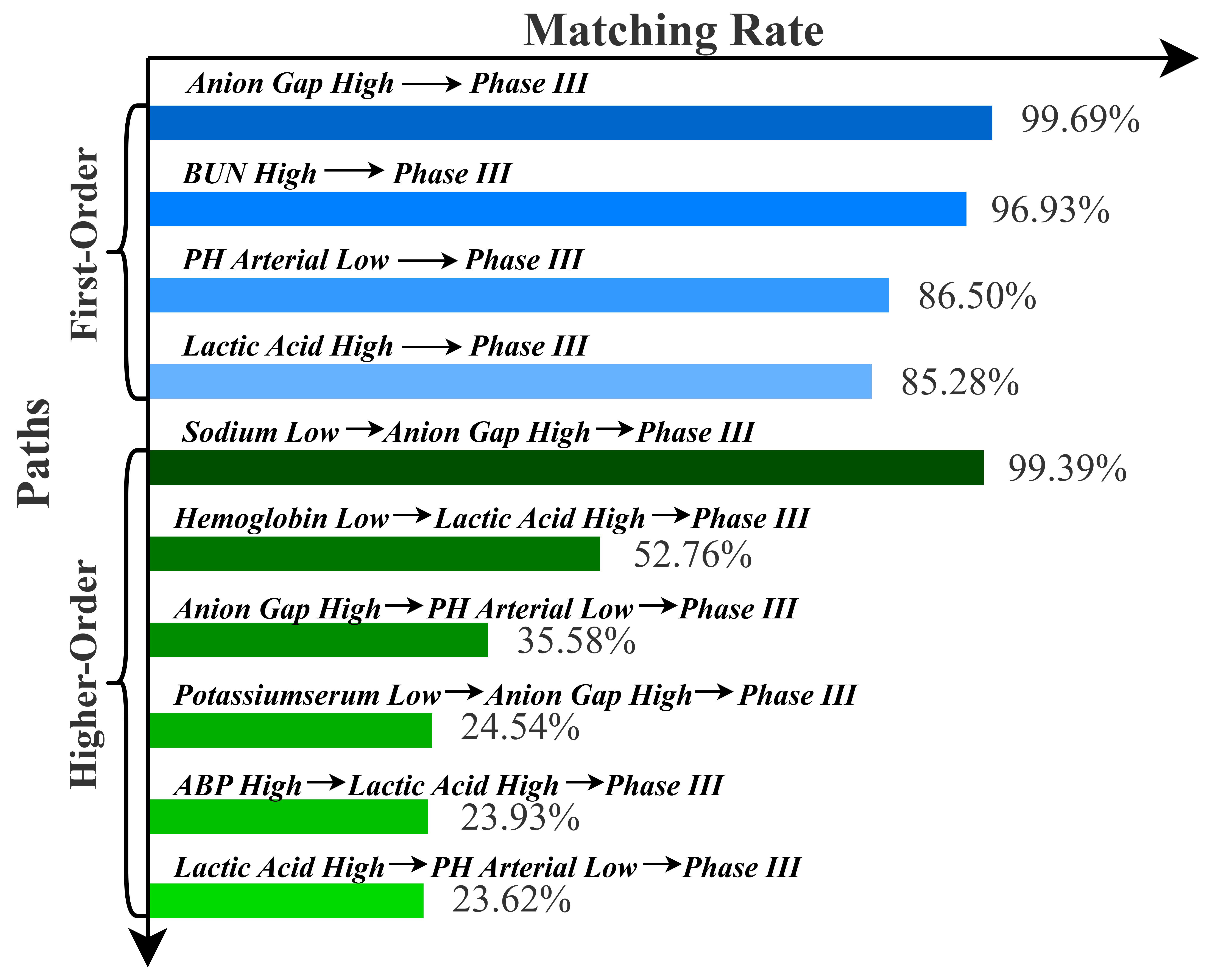}
    \caption{Matching rate of each clinical causal path.}
    \label{fig:causal-match-bar}
\end{figure}

Figure~\ref{fig:causal-match-bar} visualizes the matching rates of all paths. First, MOCHA shows excellent performance in identifying first-order direct causality. The high matching rates for Anion Gap High $\rightarrow$ Phase III (99.69\%) and BUN High $\rightarrow$ Phase III (96.93\%) indicate that the model effectively recovers direct causal relationships. The average matching rate for first-order causal relationships is 92.10\%, reflecting the model’s high accuracy in detecting direct causality. More importantly, MOCHA demonstrates substantial improvements in uncovering higher-order causal relationships. Despite the inherent challenges of multi-order causal reasoning, it achieves strong performance on several complex paths. In particular, the path Sodium Low $\rightarrow$ Anion Gap High $\rightarrow$ Phase III attains an exceptionally high matching rate of 99.39\%, even exceeding that of some first-order relationships. This performance highlights MOCHA’s advanced capability in handling multi-order causality.

These results demonstrate that MOCHA can recover ground-truth causal paths through the learned dynamic causal structures. By revealing causal origins of disease progression, MOCHA provides practical insights for temporal reasoning and diagnosis support. 

\vspace{\baselineskip}

\noindent \textbf{RQ4: Can MOCHA Reveal New Criteria?}

\begin{figure}[t]
    \centering
    \includegraphics[width=1\linewidth]{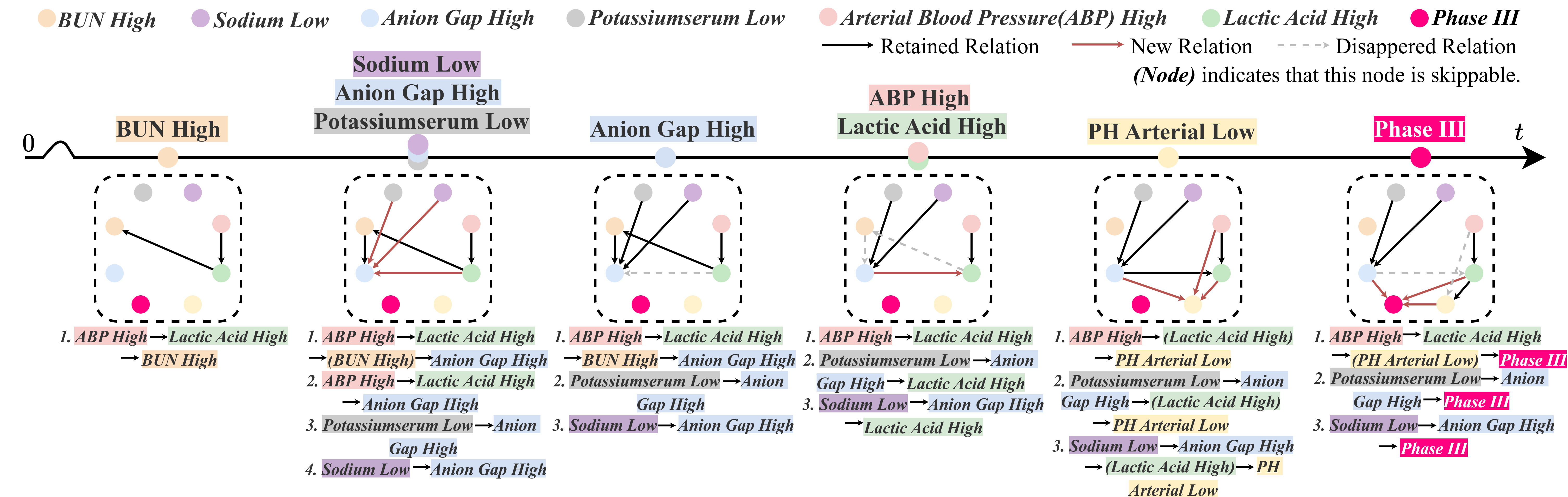}
    \caption{Dynamic causal graphs and multi-order causal paths discovered by MOCHA for an AKI patient. MOCHA captures first-, second-, and third-order causal paths across time. In the final six time steps, the model identifies increasingly complex structures. Notably, in the last three time steps, the first causal path AHP High $\rightarrow$ Lactic Acid High $\rightarrow$ PH Arterial Low $\rightarrow$ Phase III shows progressive expansion from first-order path to third-order path, illustrating how higher-order relations emerge and stabilize as the patient's condition deteriorates.}
    \label{fig:case-graph}
\end{figure}

To assess whether MOCHA can uncover novel or underexplored clinical insights, we conduct a case analysis on an AKI patient from the MIMIC-IV dataset, focusing on the final six time steps preceding Phase III onset. The learned causal graphs at each time step are visualized in Figure~\ref{fig:case-graph}, revealing how causal structures evolve as the patient’s condition deteriorates.

One representative third-order path discovered by MOCHA is: Sodium Low $\rightarrow$ Anion Gap High $\rightarrow$ Lactic Acid High $\rightarrow$ PH Arterial Low. This path reflects a clinically plausible cascade linking electrolyte imbalance, metabolic acidosis, and acid–base disturbance. MOCHA successfully identifies this indirect causal structure at multiple time steps: Lactic Acid High acts as a mediating variable that gradually bridges early abnormalities to the eventual drop in arterial pH, eventually leading to AKI Phase III. Recent clinical research confirms this mechanism: hyponatremia can result in dilutional acidosis and elevated anion gap, which in turn promotes lactic acid accumulation and subsequent acidemia \cite{kraut2014lactic}. Importantly, MOCHA identifies such multi-order causal paths without human intervention.

The emergence of such a consistent, nontrivial pattern suggests MOCHA's ability not only to recover known clinical mechanisms but also to reveal latent multi-step criteria that could inform early warning systems or risk stratification in practice.

\section{Conclusion}

In this paper, we propose MOCHA, a novel framework for modeling Multi-Order Causal  Hierarchical Architecture, which integrates dynamic causal discovery into temporal point process modeling. MOCHA introduces a learnable time decay module and a dynamic causal graph learner that captures multi-order causal dependencies over time. By formulating causal inference through a differentiable acyclicity constraint, MOCHA enables interpretable learning of causal graphs. Comprehensive experiments demonstrate that MOCHA achieves state-of-the-art performance in event likelihood estimation. Importantly, MOCHA shows strong ability in recovering validated causal paths, offering interpretable insights through dynamic graph visualizations. Beyond performance, MOCHA provides a principled way to decompose complex temporal interactions into layered causal structures and reveal new criteria.

\bibliography{references}

\end{document}